\documentclass[10pt,oneside]{article}

\usepackage[T1]{fontenc}
\usepackage[utf8]{inputenc}
\usepackage{mathptmx}
\usepackage{microtype}
\usepackage{graphicx}
\usepackage{hyperref}
\usepackage{amsmath}
\usepackage{amssymb}
\usepackage{booktabs}
\usepackage{geometry}
\usepackage{natbib}
\usepackage{titlesec}
\usepackage{tabularx}
\usepackage{array}
\usepackage[section]{placeins}  

\hypersetup{
  colorlinks=true,
  linkcolor=blue,
  filecolor=magenta,
  urlcolor=blue,
  citecolor=blue
}

\titleformat{\section}{\normalfont\fontsize{12}{14}\bfseries}{\thesection}{1em}{}
\titleformat{\subsection}{\normalfont\fontsize{11}{13}\bfseries}{\thesubsection}{1em}{}
\titleformat{\subsubsection}{\normalfont\fontsize{10}{12}\bfseries}{\thesubsubsection}{1em}{}

\counterwithin{equation}{section}
\counterwithin{figure}{section}
\counterwithin{table}{section}

\graphicspath{{media/}}

\title{GeoNDC: A Queryable Neural Data Cube for Planetary-Scale Earth Observation}

\author{
  Jianbo Qi$^{1}$\thanks{Corresponding authors: jianboqi@bnu.edu.cn (J. Qi), wangqiao@bnu.edu.cn (Q. Wang)}, 
  Mengyao Li$^{1}$, 
  Baogui Jiang$^{1}$, 
  Yidan Chen$^{1}$, 
  Xihan Mu$^{2}$,
  Qiao Wang$^{1}$\footnotemark[1]\\[4pt]
  \small $^{1}$Advanced Interdisciplinary Institute of Satellite Applications, Beijing Normal University\\
  \small Beijing 100875, China\\
  \small $^{2}$State Key Laboratory of Remote Sensing and Digital Earth, Faculty of Geographical Science\\
  \small Beijing Normal University, Beijing 100875, China
}

\date{}

\usepackage{chngcntr}

\counterwithout{figure}{section}
\counterwithout{table}{section}
\counterwithout{equation}{section}

\begin{document}

\maketitle

\begin{abstract}
Satellite Earth observation has accumulated massive spatiotemporal archives essential for monitoring environmental change, yet these remain organized as discrete raster files, making them costly to store, transmit, and query. We present GeoNDC, a queryable neural data cube that encodes planetary-scale Earth observation data as a continuous spatiotemporal implicit neural field, enabling on-demand queries and continuous-time reconstruction without full decompression. Experiments on a 20-year global MODIS MCD43A4 reflectance record ($8016 \times 4008$ pixels, 7 bands, 915 temporal frames) show that the learned representation supports direct spatiotemporal queries on consumer hardware. On Sentinel-2 imagery (10 m), continuous temporal parameterization recovers cloud-free dynamics with high fidelity ($R^2 > 0.85$) under simulated 2-km cloud occlusion. On HiGLASS biophysical products (LAI and FPAR), GeoNDC attains near-perfect accuracy ($R^2 > 0.98$). The representation compresses the 20-year MODIS archive to 0.44\,GB---approximately 95:1 relative to an optimized Int16 baseline---with high spectral fidelity (mean $R^2 > 0.98$, mean RMSE $= 0.021$). These results suggest GeoNDC offers a unified AI-native representation for planetary-scale Earth observation, complementing raw archives with a compact, analysis-ready data layer integrating query, reconstruction, and compression in a single framework.
\end{abstract}


\medskip\noindent\textbf{Subjects:} Physics - Geophysics (physics.geo-ph);Computer Science - Computer Vision and Pattern Recognition (cs.CV)

\begin{figure}[t]
    \centering
    \includegraphics[width=\linewidth]{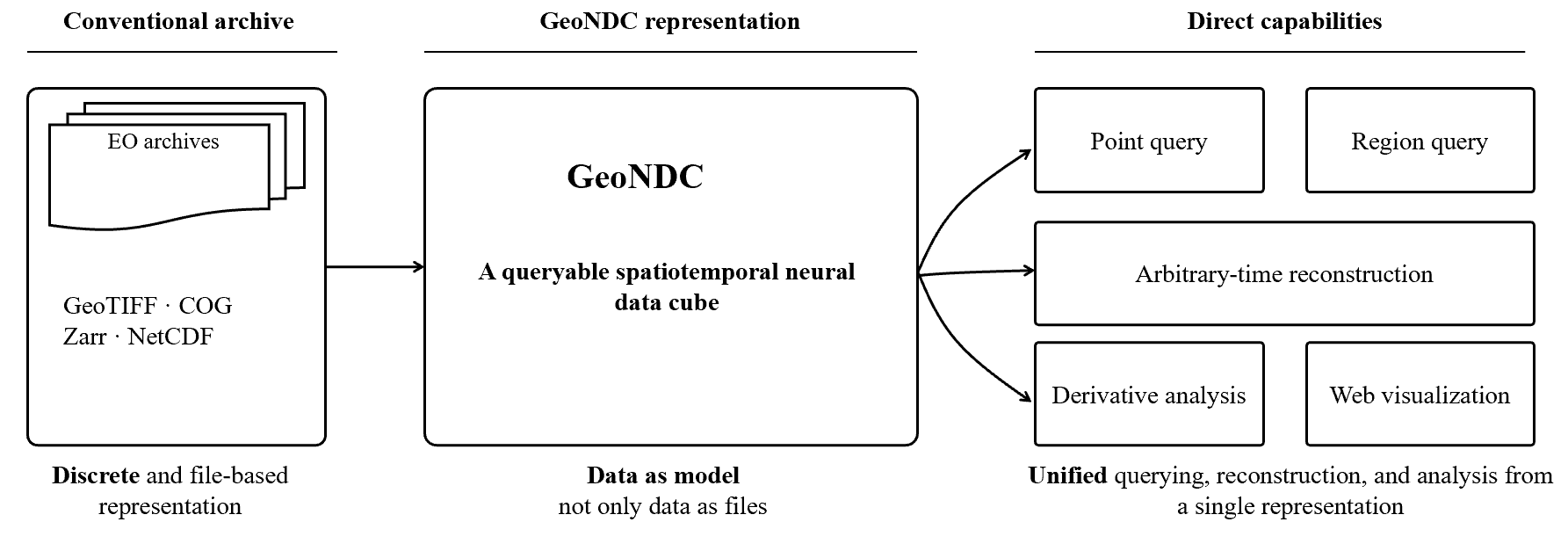}
    \caption{\textbf{Overview of the GeoNDC representation paradigm}. Conventional Earth observation archives are typically organized as discrete, file-based raster tiles or chunked arrays across space and time. GeoNDC reformulates such archives as a queryable spatiotemporal neural data cube, enabling unified point and region queries, arbitrary-time reconstruction, and web-based visualization from a single executable representation.}
    \label{fig:framework}
\end{figure}

\newpage

\section{Introduction}\label{sec:introduction}

Over the past decades, Earth observation (EO) has evolved into a foundational infrastructure for monitoring planetary change~\cite{zhu2014continuous}. Long-term satellite missions such as MODIS and Landsat have accumulated massive spatiotemporal archives, enabling the quantification of processes ranging from global greening to ecosystem resilience loss \cite{gorelick2017google,zhu2016greening,boulton2022pronounced}. Yet the central challenge of these archives is no longer data acquisition alone. At planetary scale, EO systems increasingly face a coupled bottleneck of storage, access, query, and analysis: vast data volumes must not only be preserved, but also transmitted efficiently, queried interactively across space and time, and transformed into scientifically meaningful indicators \cite{chi2016big,li2016geospatial}. As sensor constellations continue to improve revisit frequency, spatial resolution, and spectral richness, this gap between data availability and data usability is becoming a major constraint on Earth system science.

Although substantial progress has been made in cloud-native geospatial infrastructures, learned compression, and neural representations, these advances still address the problem only in partial form. Existing archive systems make EO data easier to store and distribute, while compression methods reduce file size and neural fields provide continuous signal parameterizations. However, none of these directions directly yields a planetary-scale EO representation that is simultaneously compact, geospatially queryable, and continuous in time. As a result, storage, retrieval, and reconstruction remain largely separated stages in current EO workflows.

This limitation becomes especially pronounced for long-term geospatial data cubes. Many land-surface processes are continuous in both space and time, yet satellite observations remain discrete, incomplete, and frequently interrupted by clouds, shadows, and irregular sampling. In practice, this forces downstream analysis to rely on compositing, masking, or interpolation \cite{king2013spatial}. What is missing is not simply a better codec or a faster archive format, but a representation in which compact storage, direct query, and continuity-aware reconstruction are unified by design.

Motivated by this gap, we present \emph{GeoNDC} (Geographic Neural Data Cube), a queryable neural data cube for planetary-scale Earth observation. GeoNDC represents georeferenced EO observations as a continuous spatiotemporal neural field conditioned on spatial coordinates and time, allowing the archive to be stored not merely as raster files, but as a compact executable model. In this formulation, querying a pixel value becomes function evaluation and temporal interpolation becomes native continuous-time inference. GeoNDC is therefore not merely a compression tool; it is an AI-native representation and access paradigm for EO data cubes that unifies storage, query, and reconstruction within a single framework.

Building on multi-resolution hash encodings and a decoupled spatiotemporal architecture, GeoNDC is designed to preserve the practical semantics of EO archives while changing their underlying representation. As a lossy neural representation, GeoNDC is not intended to replace authoritative raw EO archives; rather, it provides an analysis-ready and AI-ready layer for compact access and downstream scientific use. It supports geospatially aware random access, continuous reconstruction under missing observations, and compact deployment on standard hardware. This shift from \emph{data as files} to \emph{data as models} provides a new perspective on planetary EO archives: instead of distributing ever-growing collections of discrete rasters, we can learn portable neural data cubes that retain the dominant spatiotemporal structure of the observations while remaining directly queryable.

The main contributions of this study are as follows. First, we propose GeoNDC, a queryable neural data cube that reformulates planetary-scale EO archives as continuous spatiotemporal neural representations. Second, we show that this representation achieves orders-of-magnitude compression while preserving the dominant spatial and temporal dynamics of long-term satellite observations. Third, we demonstrate that the learned neural cube supports temporally coherent reconstruction under missing observations. Together, these results suggest a new route toward analysis-ready and AI-ready EO infrastructure, in which compression, access, and reconstruction are integrated into a unified representational framework.

\section{Related Work}\label{sec:related}

\subsection{Earth observation data cubes and cloud-native geospatial infrastructures}

A long-standing challenge in Earth observation is that the scientific value of satellite archives is often constrained not only by data volume, but also by the difficulty of organizing, accessing, and repeatedly analyzing large spatiotemporal collections. Earth observation data cubes (EODCs) were developed precisely to address this problem by transforming large image archives into analysis-ready, spatiotemporally aligned data structures that facilitate repeated querying and large-scale computation \cite{lewis2016eodc,lewis2017agdc,giuliani2019eoscience,appel2019gdalcubes}. Related work has further emphasized interoperability, view-based architectures, and cloud-native access patterns as essential ingredients for scalable Earth data systems \cite{nativi2017viewcube,giuliani2019interop,becker2015cloud}. 

These developments have significantly improved the practicality of large EO archives. However, they remain predominantly \emph{file-centric}: even when organized as data cubes or cloud-optimized rasters, the underlying representation is still a collection of discrete arrays to be stored, streamed, opened, and decoded before analysis. In other words, current EODC and cloud-native infrastructures substantially improve \emph{how} EO data are managed and accessed, but do not fundamentally change \emph{what} the archive is represented as. GeoNDC is motivated by this distinction. Rather than reorganizing the archive into a more efficient raster-centric structure, it seeks to represent the archive itself as a compact neural data cube that is directly queryable in space and time.

\subsection{Implicit neural representations and learned compression}

In parallel with advances in geospatial data infrastructure, computer vision and machine learning have developed a rich literature on implicit neural representations (INRs), also referred to as coordinate-based neural fields. Instead of storing a signal explicitly on a discrete grid, INRs represent it as a continuous function parameterized by a neural network. Foundational work such as NeRF \cite{mildenhall2022nerf}, SIREN \cite{sitzmann2020siren}, and instant-ngp \cite{muller2022instant} demonstrated that complex visual and geometric signals can be represented compactly, queried continuously, and differentiated analytically with respect to their input coordinates. These properties make INRs especially attractive for signals that exhibit strong spatial or spatiotemporal regularity.

A related body of work has explored learned compression and, more specifically, INR-based compression. Surveys of learning-driven image compression have documented the rapid progress of end-to-end learned codecs and the broader shift toward learned latent representations \cite{jamil2023survey}. More directly related to GeoNDC, recent work has investigated implicit neural representations as compression models, showing that continuous neural parameterizations can serve as effective lossy representations for images and other structured signals \cite{dai2025implicit, strumpler2022inrcompression}.

Nevertheless, most of this literature focuses on natural images, visual scenes, or generic signals, where the primary goal is reconstruction quality or coding efficiency. It does not directly address a core requirement of Earth observation archives: the need for a geospatially aware, temporally indexed representation that supports not only compression, but also point-wise query, regional access, continuous-time reconstruction, and downstream scientific analysis. GeoNDC builds on the representational advantages of INRs, but applies them to a different object: the planetary-scale EO data cube.

To clarify the position of GeoNDC relative to existing geospatial storage formats, array-based data cube systems, conventional and learned compression methods, and general implicit neural representations, Table \ref{tab:positioning_compact} summarizes the key differences along several dimensions relevant to Earth observation archives and scientific analysis.

\subsection{Neural representations in Earth observation and geoscience}

Neural fields and implicit representations have recently been extended into remote sensing and geoscience, especially for 3D reconstruction, satellite photogrammetry, and scientific field modeling. In remote sensing, NeRF-like approaches have been adapted to multi-view satellite imagery for novel-view synthesis, elevation prediction, and neural surface reconstruction \cite{li2023rsinrcompress, xie2023rsnerf,qu2023satmesh}. In geoscience, GeoINR has shown that implicit neural representations can provide flexible and scalable models for three-dimensional geological structures \cite{hillier2023geoinr} and Earth systems model data \cite{mostajeran2025context}. These studies establish that continuous neural representations are compatible with geospatial signals and can be successfully adapted to Earth-related domains.

However, the targets of these studies are fundamentally different from the focus of GeoNDC. Most existing EO neural-field work is designed for local 3D scene reconstruction, photogrammetric geometry, or object-level representation, rather than for long-term, georeferenced, planetary-scale spatiotemporal archives. Even when these approaches operate on Earth observation imagery, the representation is typically optimized for view synthesis or geometric inference, not for compact storage and direct access to a global EO data cube. GeoNDC therefore occupies a distinct position: it treats the archive itself—rather than a local 3D scene or a reconstructed surface—as the object of neural parameterization, enabling it to be directly queried, reconstructed, and analyzed as a continuous spatiotemporal function.

\subsection{Reconstruction under incomplete observations}

A separate and highly relevant line of work concerns the reconstruction of cloud-affected or otherwise incomplete remote sensing time series. Existing approaches include gap-filling, interpolation, similar-pixel replacement, tensor or low-rank regularization, and climate-informed temporal reconstruction \cite{weiss2014gapfill,cheng2014cloudremoval,duan2020thickcloud,yu2021cgf}. These methods have shown that continuity assumptions and auxiliary structure can substantially improve the usability of EO time series under missing observations. However, they generally treat reconstruction as a dedicated preprocessing stage applied \emph{before} downstream analysis, rather than as an intrinsic property of the representation itself.

This distinction matters because many scientific questions require more than cloud-free snapshots. They require a representation that is temporally coherent and queryable. Existing data-cube infrastructures improve access to analysis-ready rasters, and existing gap-filling methods improve observation completeness, but neither class of methods natively turns the EO archive into a continuous spatiotemporal object. GeoNDC is designed to bridge this gap. By fitting a continuous neural field directly to incomplete EO observations, it integrates compact representation, continuity-aware recovery, and direct query into a single framework.

Taken together, the literature suggests that the ingredients required by GeoNDC already exist in partial form: EODCs provide scalable archive organization, cloud-native formats improve partial access, INRs provide continuous representations, and remote sensing reconstruction methods address missing observations. The novelty of GeoNDC lies in combining these ideas into a unified \emph{queryable neural data cube} for planetary-scale Earth observation. Rather than viewing EO archives only as files to be stored, streamed, and post-processed, GeoNDC treats them as compact neural representations that can be directly queried and reconstructed.

\begin{table*}[t]
\centering
\caption{Conceptual positioning of GeoNDC relative to existing geospatial storage, compression, and neural representation paradigms.}
\label{tab:positioning_compact}
\renewcommand{\arraystretch}{1.15}
\setlength{\tabcolsep}{4pt}
\begin{tabularx}{\textwidth}{>{\raggedright\arraybackslash}p{2.8cm} *{5}{>{\centering\arraybackslash}X}}
\hline
\textbf{Method} & \textbf{Native geospatial semantics} & \textbf{Unified spatiotemporal representation} & \textbf{Continuous query support} & \textbf{Selective access} & \textbf{Analysis-oriented design} \\
\hline
GeoTIFF / COG               & Yes     & Partial & No  & Yes     & Yes \\
Zarr / NetCDF data cubes    & Yes     & Yes     & No  & Yes     & Yes \\
Conventional compression    & No      & Partial & No  & Limited & No \\
Learned image compression   & No      & No      & No  & Limited & No \\
General INR / neural fields & No      & Partial & Yes & Yes     & Partial \\
\textbf{GeoNDC}             & \textbf{Yes} & \textbf{Yes} & \textbf{Yes} & \textbf{Yes} & \textbf{Yes} \\
\hline
\end{tabularx}
\end{table*}

\section{The Continuous GeoNDC Framework}\label{sec:framework}

\subsection{Mathematical Formulation}\label{sec:mathematical}
Let a georeferenced EO archive be represented as a spatiotemporal data cube
\begin{equation}
\mathbf{D} \in \mathbb{R}^{H \times W \times T \times C},
\end{equation}
where $H$ and $W$ denote the spatial dimensions, $T$ is the number of temporal observations, and $C$ is the number of physical variables or spectral channels. In conventional storage systems, $\mathbf{D}$ is materialized explicitly as a large collection of raster arrays. GeoNDC instead seeks a compact continuous representation
\begin{equation}
\Phi_{\theta}: \mathcal{M} \times \mathcal{T} \rightarrow \mathbb{R}^{C},
\end{equation}
where $\mathcal{M}$ denotes the spatial manifold and $\mathcal{T}$ denotes the temporal domain. Given a query coordinate $(x,y,t)$, the network predicts the corresponding surface state
\begin{equation}
\hat{\mathbf{v}} = \Phi_{\theta}(x,y,t),
\end{equation}
where $\hat{\mathbf{v}} \in \mathbb{R}^{C}$ is the reconstructed vector of physical values at that spacetime location.

This reformulation changes the EO archive from an explicitly stored tensor into a learned function parameterized by $\theta$. As a result, the storage cost no longer scales directly with $H \times W \times T \times C$, but with the number of trainable parameters and any optional correction terms. More importantly, the archive becomes directly queryable at arbitrary spacetime coordinates. Point-wise query, regional reconstruction, and temporal interpolation can all be performed by evaluating the same continuous function.

Because optical satellite observations are frequently affected by clouds, shadows, and sensor artifacts, not all samples in $\mathbf{D}$ are valid. Let $\mathbf{M} \in \{0,1\}^{H \times W \times T}$ denote the validity mask, where $M(x,y,t)=1$ indicates a valid observation and $M(x,y,t)=0$ indicates contamination or missing data. GeoNDC is trained only on valid observations by minimizing a masked objective:
\begin{equation}\label{eq:maskedlearning}
\mathcal{L}_{\mathrm{rec}} =
\frac{\sum_{x,y,t} M(x,y,t)\,\|\Phi_{\theta}(x,y,t)-\mathbf{v}(x,y,t)\|_2^2}
{\sum_{x,y,t} M(x,y,t)}.
\end{equation}
This formulation allows the network to learn from incomplete EO records without forcing it to fit contaminated pixels. Missing observations are then recovered through the continuity of the learned spatiotemporal field rather than through an external gap-filling procedure. Because $M(x,y,t)\in\{0,1\}$, invalid or contaminated observations do not contribute to the numerator, while the denominator counts the number of valid samples.

\subsection{The GeoNDC Architecture}\label{sec:architecture}
The key design challenge is that EO data cubes are highly anisotropic in space and time. Spatial patterns often contain sharp boundaries and high-frequency structures, such as coastlines, cropland parcels, and urban edges, whereas temporal evolution is typically smoother and more correlated, especially for vegetation and biophysical variables. To capture this asymmetry efficiently, GeoNDC adopts a decoupled dual-branch architecture.

\begin{figure}[htbp]
\centering
\includegraphics[width=0.8\textwidth]{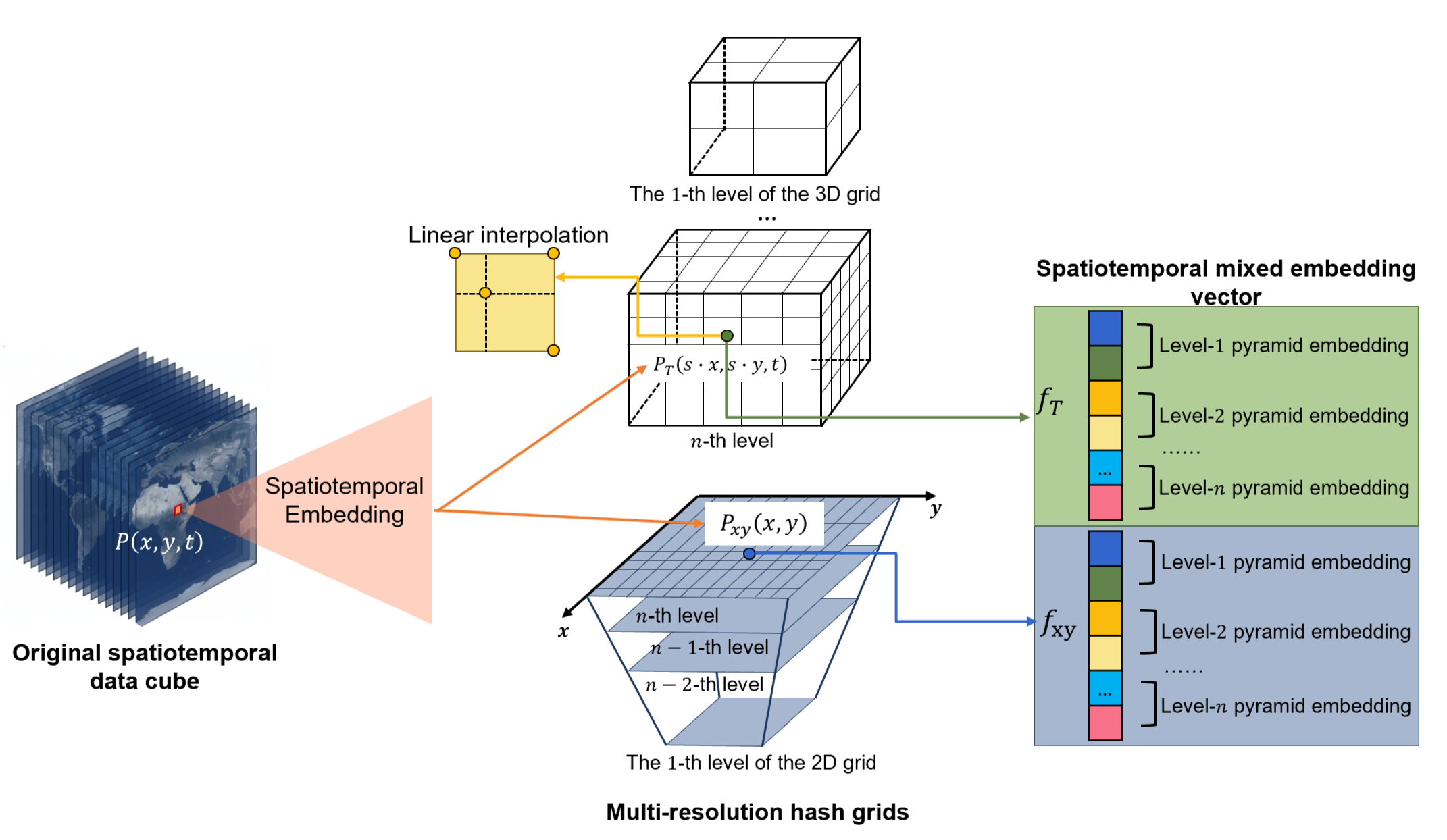}
\caption{\textbf{Architecture of the spatiotemporal embedding.} 
A query point $(x,y,t)$ from the original spatiotemporal data cube is processed through a decoupled dual-branch structure designed to address the intrinsic spatiotemporal anisotropy of Earth observations. To preserve high-frequency spatial boundaries, the spatial coordinates $(x,y)$ are queried directly in a static high-resolution 2D HashGrid (bottom). To capture smoother regional spatiotemporal dynamics while reducing temporal striping artifacts, a spatial scaling factor $s$ is applied to generate scaled coordinates $(s x, s y, t)$, which are then fed into a coarse 3D HashGrid (top). The hierarchical descriptors extracted by interpolation, $\mathbf{f}_{xy}$ and $\mathbf{f}_{T}$, are concatenated to form the final spatiotemporal mixed embedding vector.}\label{fig:embedding}
\end{figure}

As shown in Fig.\ref{fig:embedding}, this architecture addresses the intrinsic spatiotemporal anisotropy of Earth observations through a factorized design with a Static High-Resolution 2D Branch and a Dynamic Coarse 3D Phenology Branch:

\textbf{Static High-Resolution 2D Branch.} This branch is designed to lock the relative invariant physical boundaries of ground objects. It takes the normalized geographic coordinates $(x, y)$ as input and queries a high-capacity, multi-resolution 2D HashGrid \cite{muller2022instant}. By allocating a deep resolution pyramid exclusively to the spatial dimensions, this static embedding ensures that the sharp geometric textures of the landscape are preserved across the entire decadal time series without temporal degradation.

\textbf{Dynamic Coarse 3D Temporal Branch.} The second branch models the continuous temporal evolution of the land surface. Crucially, instead of feeding the original high-resolution coordinates directly into the temporal network, we introduce a Spatial Downscaling Mechanism. We apply a spatial scaling factor ($s < 1$) to the spatial dimensions, generating scaled coordinates $(s \cdot x, s \cdot y, t)$ to query a multi-resolution 3D HashGrid. This spatial scaling forces the 3D grid to operate at a broader receptive field, effectively reducing spatial redundancy and suppressing temporal striping artifacts. Consequently, the network is constrained to learn smooth, regional temporal trends--- mimicking the natural progression of large-scale phenological dynamics while preventing temporal overfitting.

For a query point $P(x,y,t)$, the spatial coordinates $(x,y)$ are first processed by the high-resolution static 2D HashGrid encoder to preserve fine spatial details:
\begin{equation}
\mathbf{f}_{xy} = E_{xy}(x,y).
\end{equation}

In parallel, the coarse 3D spatiotemporal branch is used to model temporally smooth regional dynamics:
\begin{equation}
\mathbf{f}_{T} = E_{T}(s x, s y, t).
\end{equation}

\begin{figure}[htbp]
\centering
\includegraphics[width=0.8\textwidth]{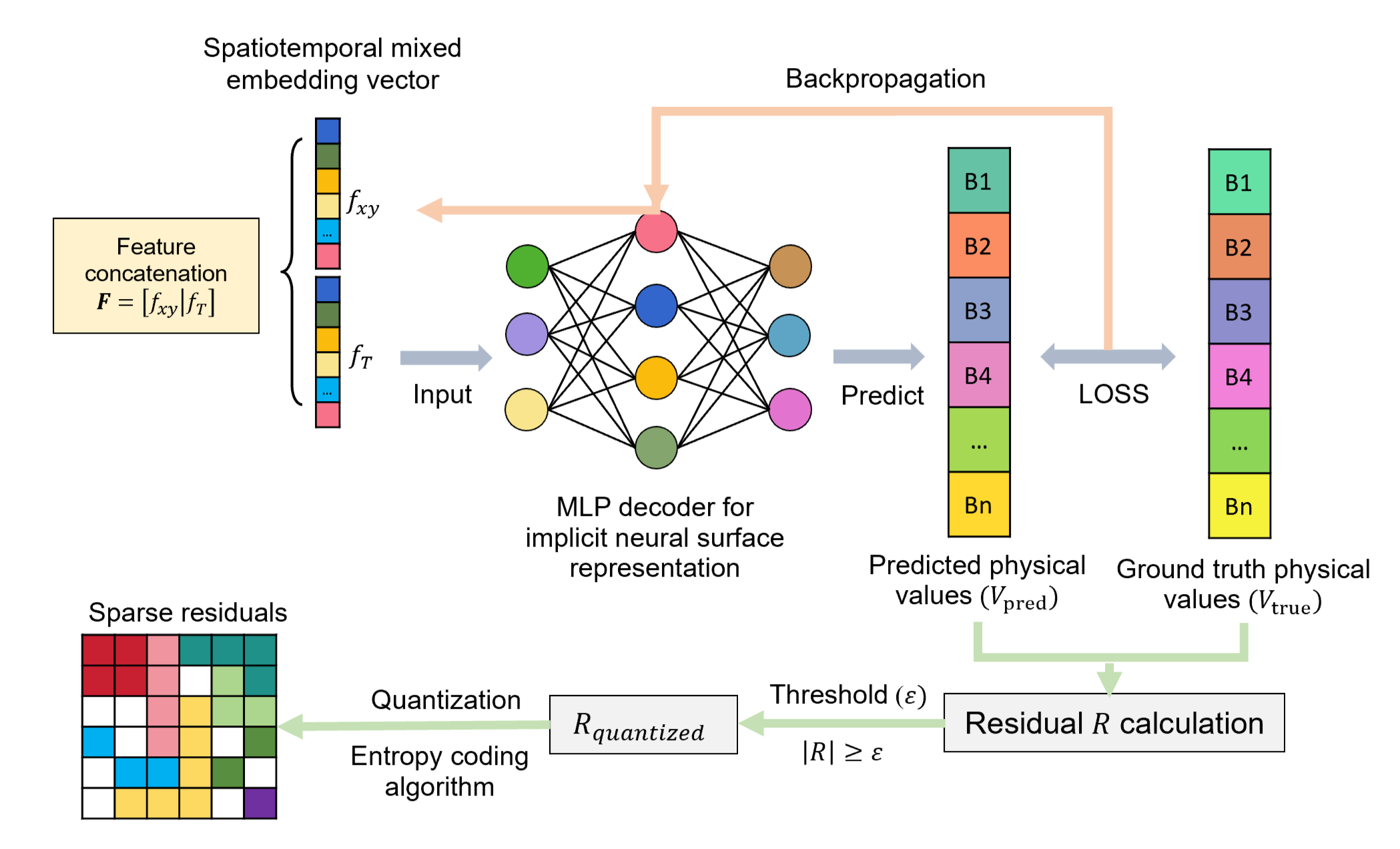}
\caption{\textbf{The GeoNDC architecture.} 
The framework maps coordinates from the original spatiotemporal data cube into a high-dimensional feature space through spatiotemporal geometric embedding. The resulting mixed embedding vector $\mathbf{F}$ is decoded by an MLP to predict the surface state $\widehat{\mathbf{v}}$. Model parameters are optimized through backpropagation by comparing the prediction $\widehat{\mathbf{v}}$ with the ground-truth value $\mathbf{v}$. To preserve localized high-frequency details, residuals $\mathbf{r} = \mathbf{v} - \widehat{\mathbf{v}}$ whose magnitude exceeds a threshold $\tau$ are quantized and stored in a sparse residual package using entropy coding, forming an optional correction layer on top of the base neural representation.}\label{fig:architecture}
\end{figure}

Using the spatiotemporal embedding described above, a spatiotemporal coordinate $(x, y, t)$ can be described as a learnable high-dimensional vector $\mathbf{F} = [\mathbf{f}_{xy};\mathbf{f}_{T}]$. This vector is then processed by an MLP decoder for implicit neural representation, which predicts the instantaneous surface state:

\begin{equation}
\widehat{\mathbf{v}} = \Psi(\mathbf{F}; \mathbf{\theta}_{\text{MLP}})
\end{equation}

where $\Psi$ denotes the MLP decoder and $\mathbf{\theta}_{\text{MLP}}$ represents its trainable parameters. As shown in Fig. \ref{fig:architecture}, this architecture acts as a soft compression algorithm. By sharing the heavy spatial parameters across the entire time series, the model effectively amortizes the storage cost of the planetary surface dynamics. This decoupling allows the representation of decadal sequences within a remarkably compact model size.

During training, the model is optimized through backpropagation by minimizing the discrepancy between the predicted physical values $\hat{\mathbf{v}}$ and the ground-truth physical values $\mathbf{v}$. To further enhance reconstruction fidelity in cases where highly localized deviations remain after neural approximation, we optionally introduce a sparse residual layer. Let
\begin{equation}
\mathbf{r}(x,y,t) = \mathbf{v}(x,y,t) - \hat{\mathbf{v}}(x,y,t)
\end{equation}
denote the reconstruction residual. For regions where the absolute residual $|\mathbf{r}|$ exceeds a predefined threshold $\tau$, the residual values are quantized and encoded into a sparse residual package using entropy coding. This dual-layer strategy, combining the base neural representation with an optional sparse residual correction, allows GeoNDC to maintain high compression while preserving critical local spectral details.

\FloatBarrier
\subsection{Learning from Incomplete Observations}

A major challenge in optical EO is the pervasive loss of valid observations due to clouds, haze, shadows, and sensor artifacts. Traditional reconstruction workflows often rely on Maximum Value Compositing (MVC) or temporal interpolation as external preprocessing steps. While useful in some cases, these operations are separate from the data representation itself and may introduce temporal discontinuities or spectral inconsistencies when observations from different dates are merged.

In GeoNDC, missing-data recovery is handled natively during model fitting through the masked objective in Eq.~\ref{eq:maskedlearning}. The network is trained strictly on valid observations and never forced to reproduce contaminated values. Because the learned parameters are shared over the continuous temporal domain, the model is implicitly constrained to organize the available observations into a coherent spatiotemporal manifold. When the model encounters a cloud-masked or missing region at time $t$, it cannot rely on a direct observation at that location. Instead, it estimates the value from the learned temporal context and neighboring valid observations that jointly shape the continuous field.

The key advantage of this approach lies in the interaction between the masked optimization and the factorized spatiotemporal architecture in Section \ref{sec:architecture}. Missing-value recovery is therefore not implemented as a separate post-processing operation, but emerges naturally from fitting a continuous spatiotemporal representation to incomplete EO records. As a result, the reconstructed values are not merely spectrally plausible, but remain consistent with the broader temporal evolution learned from valid observations. This makes GeoNDC particularly suitable as an analysis-ready layer for applications that require temporally coherent surface dynamics rather than isolated cloud-free mosaics.

\FloatBarrier
\subsection{The GeoNDC Unified Storage Protocol}

\begin{figure}[htbp]
\centering
\includegraphics[width=0.8\textwidth]{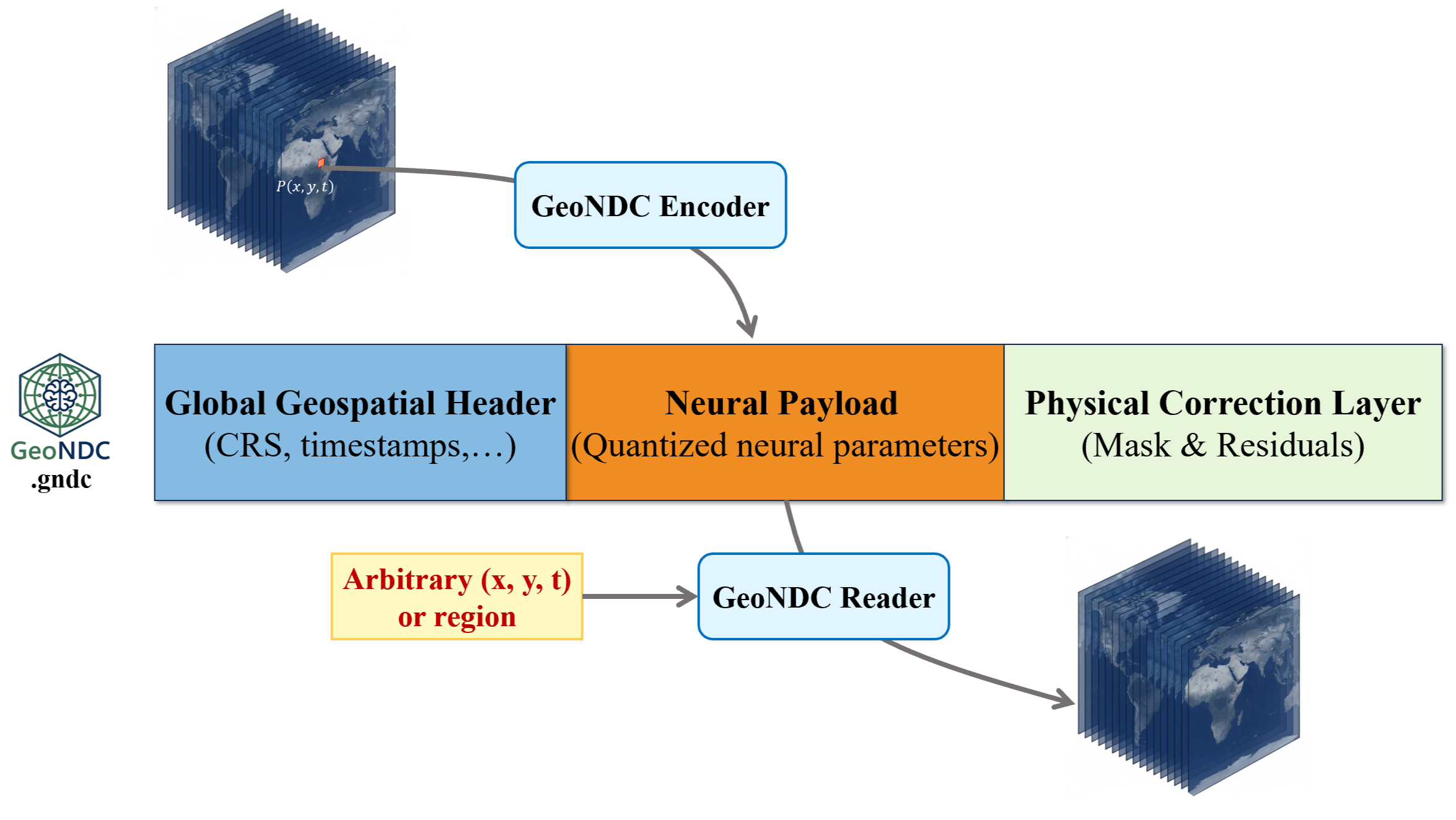}
\caption{\textbf{The GeoNDC unified storage protocol and access pipeline.} 
The GeoNDC Encoder compresses raw spatiotemporal data cubes into a compact \texttt{.gndc} file composed of three components: (1) a \textbf{Global Geospatial Header} that preserves geospatial metadata required for interoperability, including the coordinate reference system (CRS) and temporal indexing information; (2) a \textbf{Neural Payload} containing the quantized parameters of the learned neural field; and (3) an optional \textbf{Physical Correction Layer} that stores validity masks and sparse residuals. The GeoNDC Reader supports direct random access, allowing arbitrary points or regions in space and time, $(x,y,t)$, to be queried without full-volume decompression.}\label{fig:storage}
\end{figure}

To support practical distribution and interactive access, we define the GeoNDC Unified Storage Protocol, a compact binary serialization format for neural EO data cubes. Unlike traditional raster formats such as GeoTIFF, which store explicit pixel arrays, a \texttt{.gndc} file stores the parameters and metadata required to reconstruct the continuous spatiotemporal field. The goal is not to replace authoritative raw archives, but to provide a portable representation layer that can be efficiently transmitted, queried, and analyzed. As illustrated in Fig.~\ref{fig:storage}, the protocol consists of three components:

\begin{enumerate}
    \item \textbf{Global Geospatial Header.}  
    This lightweight metadata header preserves spatial awareness and interoperability with standard geospatial workflows. It stores the coordinate reference system (CRS), the spatiotemporal bounding box $\Omega$, temporal indexing information, and normalization parameters required to map geographic queries into the internal coordinate system of the neural model.

    \item \textbf{Neural Payload.}  
    The core of the \texttt{.gndc} file is the neural payload, which contains the learned parameters of the HashGrid encoders and the MLP decoder. To reduce storage overhead, these parameters can be stored in half precision (\texttt{float16}) or in quantized form when appropriate. This payload represents the base continuous field used for spatiotemporal reconstruction and query.

    \item \textbf{Physical Correction Layer (optional).}  
    To preserve the distinction between raw observations and neural reconstruction, and to retain localized high-frequency details when necessary, GeoNDC optionally stores:
    \begin{enumerate}
        \item a \textbf{Validity Bitmask}, encoded as a compressed bitstream, indicating whether each spacetime sample in the source data was originally valid or contaminated; and
        \item a \textbf{Sparse Residual Package}, which stores quantized residuals only at locations where the neural approximation exceeds a threshold.
    \end{enumerate}
\end{enumerate}

This design provides two practical advantages. First, it preserves scientific transparency by allowing users to distinguish between directly observed values and values reconstructed from the neural representation. Second, it supports on-demand random query access without requiring full decompression of the original volume. Given a spatial or spatiotemporal region of interest, the GeoNDC runtime evaluates the neural model only at the queried coordinates, making the archive executable rather than purely static. This property is central to the notion of a \emph{queryable neural data cube}.

\subsection{Implementations}\label{sec:implementation}

We implemented GeoNDC using PyTorch with Tiny CUDA Neural Networks for efficient multiresolution hash encoding. All experiments were conducted on a single workstation equipped with an NVIDIA RTX 4080 GPU (16 GB VRAM). To efficiently capture the multi-scale characteristics of long-term surface dynamics within the memory constraints of consumer hardware, we adopted an anisotropic configuration strategy. The spatial branch uses a high-resolution 2D HashGrid to preserve fine geographic structure, while the temporal branch uses a lower-resolution 3D HashGrid to encode long-term dynamics with reduced memory cost. Specifically, a significantly larger hash table capacity is allocated to the 2D spatial branch than to the coarse 3D branch, reflecting the fact that land-surface geometry typically exhibits higher-frequency variation in space than in time.

The MLP decoder consists of lightweight fully connected layers chosen to balance expressive power and query speed. In practice, the overall model capacity is controlled by the number of HashGrid levels, feature dimensions, and hash table size, allowing the framework to adapt to EO data cubes with different spatial scales and temporal lengths. During optimization, valid observations are sampled from the archive and normalized into the coordinate system of the neural field. Training is performed using standard stochastic gradient-based optimization under the masked reconstruction objective in Eq.~\ref{eq:maskedlearning}. After convergence, the learned parameters are serialized into the \texttt{.gndc} format together with the metadata header and any optional correction layers, enabling point query, regional reconstruction, and temporal interpolation on standard computing hardware. 

In addition to the native GPU runtime, we implemented a browser-based viewer that performs full neural inference on the client side using WebGPU compute shaders. The viewer loads the \texttt{.gndc} model weights into GPU storage buffers and executes hash grid lookups and MLP decoding entirely within the browser, requiring no backend server or data streaming beyond the initial model download. This enables interactive exploration of compressed EO data cubes on any WebGPU-capable device, including standard laptops and mobile browsers, without local software installation.

\section{The Evaluation of GeoNDC}\label{sec:evaluation}

\subsection{Datasets and Experimental Setup}\label{sec:datasets}

To evaluate GeoNDC across distinct spatial scales, temporal extents, and product types, we designed three complementary experiments, ranging from high-resolution local reconstruction to planetary-scale representation and analysis-ready product serving. Rather than relying on a single benchmark, this section assesses whether GeoNDC can function as a unified neural data cube under markedly different Earth observation scenarios. In particular, we examine the framework from four perspectives: reconstruction fidelity, robustness to incomplete observations, storage efficiency and deployability, and support for downstream spatiotemporal analysis. Specifically, we selected three representative datasets:

\begin{enumerate}
    \item \textbf{Sentinel-2 Time Series}: We selected a $50\,\mathrm{km} \times 50\,\mathrm{km}$ tile over Beijing, China, spanning from June 10 to July 20, 2024. With a spatial resolution of $10\,\mathrm{m}/20\,\mathrm{m}$ and 10 spectral bands, this dataset serves as the benchmark for evaluating high-resolution reconstruction fidelity, preservation of high-frequency texture, and recovery under incomplete observations. The scene contains a heterogeneous mixture of urban structures, transportation corridors, cultivated fields, and peri-urban vegetation, making it well suited for testing whether GeoNDC can preserve sharp boundaries while maintaining temporal coherence.

    \item \textbf{Global MODIS Reflectance}: We employed the MCD43A4 Nadir BRDF-Adjusted Reflectance product at global scale for the period 2005--2024. The data were resampled to a $5\,\mathrm{km}$ spatial grid ($8016 \times 4008$ pixels) and organized into a 20-year spatiotemporal cube with 7 spectral bands and 915 temporal snapshots. This dataset represents the planetary-scale scenario and was used to evaluate storage efficiency and spatiotemporal queryability on a multi-decadal Earth observation archive.

    \item \textbf{HiGLASS LAI/FPAR Products}: To demonstrate the use of GeoNDC beyond reflectance archives and toward scientific data serving, we used the high-resolution ($20\,\mathrm{m}$) HiGLASS Leaf Area Index (LAI) and Fraction of Photosynthetically Active Radiation (FPAR) products. This dataset was selected to test whether GeoNDC can compactly represent and distribute quantitative inversion products while preserving inter-variable consistency and high reconstruction fidelity.
\end{enumerate}

These three datasets were chosen to reflect distinct but complementary demands. The Sentinel-2 experiment emphasizes high-frequency spatial detail and continuity-aware reconstruction under missing observations. The global MODIS experiment evaluates whether the framework remains compact, directly queryable, and analytically useful at planetary scale. The HiGLASS experiment tests whether GeoNDC can serve as a shared neural representation for analysis-ready ecological products. Unless otherwise specified, quantitative reconstruction quality is reported using the coefficient of determination ($R^2$) and root mean square error (RMSE), while storage efficiency is quantified by comparing the raw raster archive with the resulting \texttt{.gndc} model.

\FloatBarrier
\subsection{High-Resolution Surface Reconstruction and Implicit Inpainting Performance}\label{sec:highres}

To evaluate the reconstruction fidelity and gap-recovery capability of GeoNDC at high spatial resolution, we conducted a Mask-and-Restore experiment using the Sentinel-2 time series. A cloud-free snapshot from June 25, 2024 was selected as the reconstruction target, and a custom masking protocol was applied to simulate observation gaps at multiple scales. The simulation included 50 small gaps ($100$--$200\,\mathrm{m}$, representing cumulus-scale cloud occlusions), 10 medium gaps ($500$--$800\,\mathrm{m}$, representing typical cloud patches), and 3 large gaps ($1.5$--$2\,\mathrm{km}$, representing massive cloud banks). The GeoNDC model was trained on the complete sequence from June 10 to July 20, 2024, thereby allowing the network to exploit both local spatial structure and the surrounding temporal manifold.

\begin{figure}[htbp]
\centering
\includegraphics[width=\textwidth]{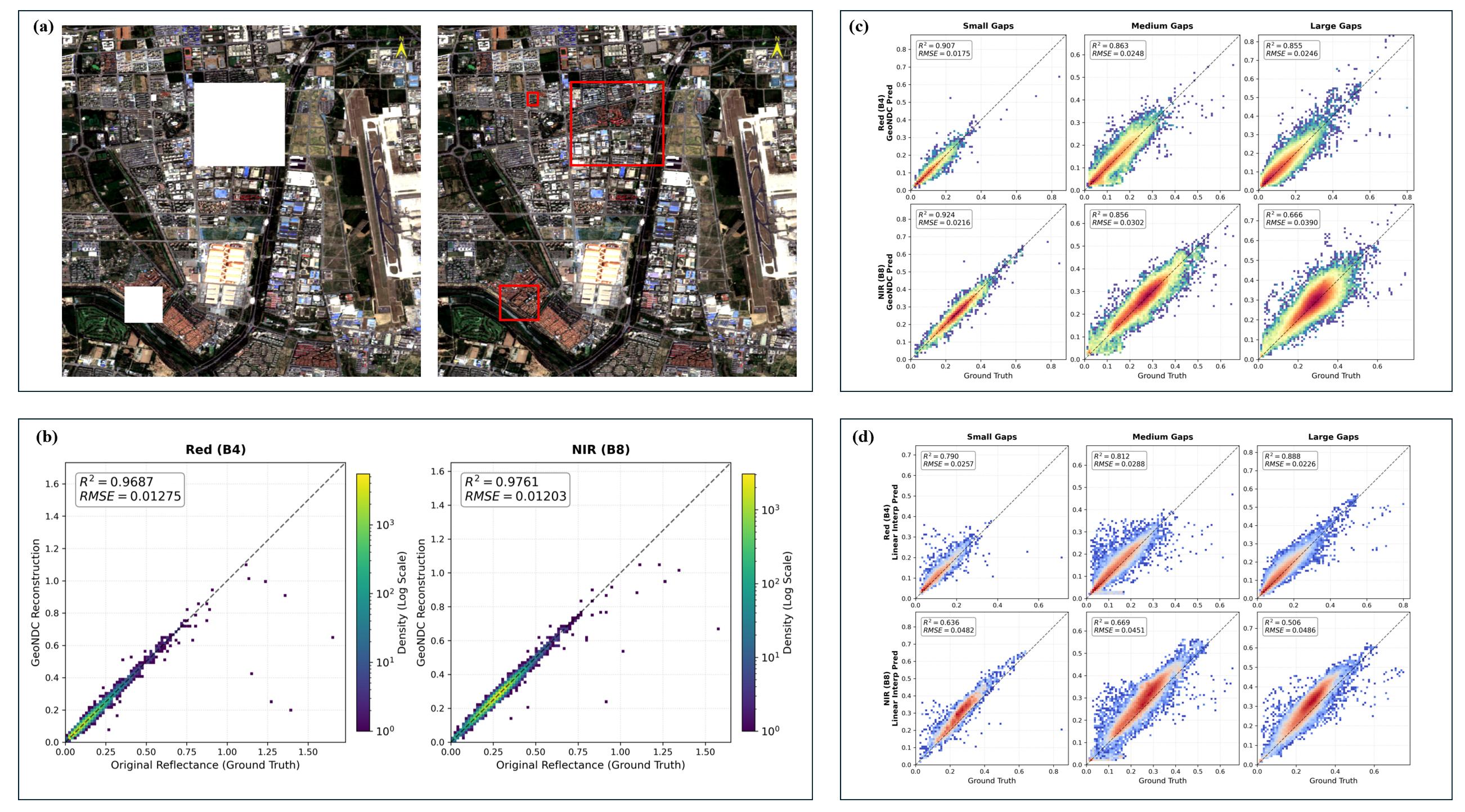}
\caption{\textbf{High-resolution surface reconstruction under incomplete observations.} 
(a) \textbf{Mask-and-Restore experiment:} Visual comparison between Sentinel-2 snapshots with simulated gaps of varying scales (small, medium, and large; left) and the corresponding GeoNDC reconstructions (right). 
(b) \textbf{Reconstruction fidelity in valid regions:} Density scatter plots for the Red (B4) and NIR (B8) bands, comparing GeoNDC reconstructions with the original observations in non-masked regions. 
(c) \textbf{Recovery accuracy across gap sizes:} Quantitative comparison between GeoNDC reconstructions and the ground-truth observations within masked regions for different gap scales. 
(d) \textbf{Interpolation baseline comparison:} Reconstruction performance of a traditional linear interpolation method over the same masked regions and spectral bands.}\label{fig:reconstruction}
\end{figure}

The results demonstrate that GeoNDC achieves high reconstruction fidelity even in complex urban--agricultural environments. In non-masked regions, the reconstructed imagery maintains strong spectral agreement with the original observations, achieving an $R^2$ of 0.9687 and an RMSE of 0.01275 for the Red (B4) band, and an $R^2$ of 0.9761 with an RMSE of 0.01203 for the NIR (B8) band, as shown in Fig.~\ref{fig:reconstruction}(a--b). This level of agreement indicates that the representation does not merely preserve broad land-cover composition, but also maintains high-frequency spatial boundaries such as urban edges, road networks, and intricate field parcels. In terms of storage efficiency, the original 4.2\,GB Sentinel-2 spatiotemporal volume is compressed into a 292\,MB \texttt{.gndc} model file, confirming that GeoNDC can transform a large stack of discrete raster observations into a compact and executable representation without losing the dominant spatial structure.

A more demanding test concerns performance within masked regions, where the model cannot rely on direct observations. Here, successful recovery depends on whether the learned neural data cube can organize valid observations into a coherent spatiotemporal manifold and infer missing surface states from this latent continuity. As shown in Fig.~\ref{fig:reconstruction}(c), the values reconstructed by GeoNDC remain strongly correlated with the original ground truth across all gap sizes, with $R^2$ values consistently above 0.85 for both spectral bands. For example, in the Red band, the model achieves $R^2 = 0.923$ for small gaps and $R^2 = 0.867$ for large gaps. This result is notable because the largest masks substantially exceed the scale of many local textural elements and therefore cannot be restored by local neighborhood statistics alone.

To place this result in context, we compared GeoNDC against a traditional temporal linear interpolation baseline over the same masked regions. As shown in Fig.~\ref{fig:reconstruction}(d), linear interpolation yields consistently lower accuracy, with the $R^2$ for the NIR band dropping to 0.636 even for small gaps. Beyond the numerical degradation, the interpolated NIR values also exhibit noticeable spectral bias across all gap scales. This is especially pronounced during the peak summer growing period, when crop development is highly non-linear and simple interpolation fails to capture biological acceleration. By contrast, GeoNDC benefits from its continuous temporal parameterization and shared spatiotemporal embedding, allowing it to infer missing values in a way that remains more consistent with the underlying phenological progression learned from valid observations on surrounding dates.

Taken together, the Sentinel-2 experiment demonstrates two important properties of GeoNDC. First, the framework preserves high-frequency spatial structure in high-resolution optical imagery. Second, when observations are incomplete, the same representation supports continuity-aware reconstruction without the severe spectral distortions observed in simple interpolation baselines. This makes GeoNDC particularly suitable as a compact representation layer for analysis-ready EO products in settings where cloud contamination and temporal incompleteness are unavoidable.

\FloatBarrier
\subsection{Planetary-scale Scalability and Queryability}\label{sec:planetary}

\begin{figure}[htbp]
\centering
\includegraphics[width=\textwidth]{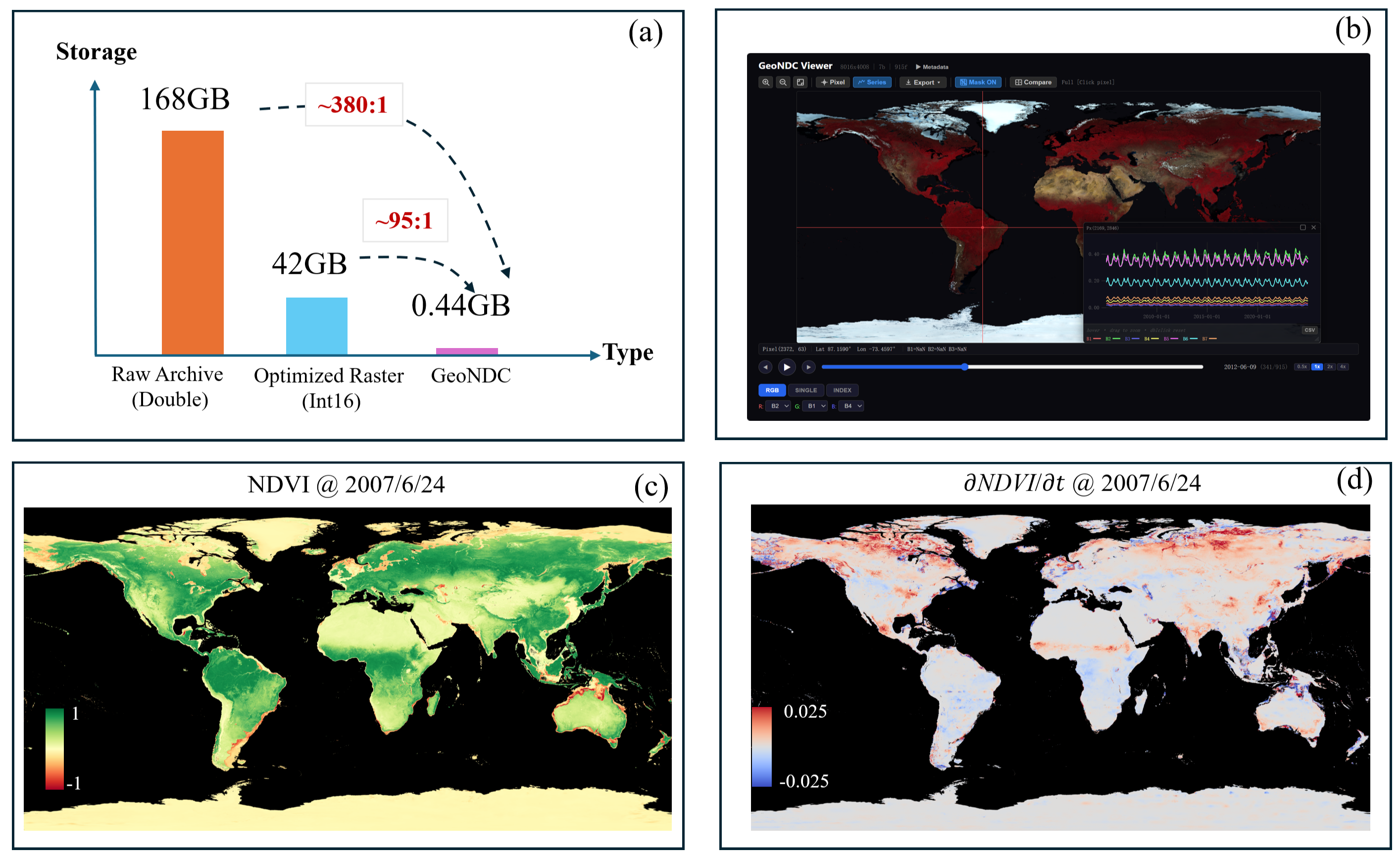}
\caption{\textbf{Planetary-scale compression and queryability.}
(a) \textbf{Storage reduction at planetary scale:} Storage comparison for the 20-year global MODIS archive. GeoNDC (0.44\,GB) achieves an approximately $380\!:\!1$ compression ratio relative to the raw double-precision archive (168\,GB), and an approximately $95\!:\!1$ ratio relative to a theoretical Int16 raster baseline (42\,GB).
(b) \textbf{Interactive query and visualization:} The compressed model supports direct spatiotemporal access on consumer hardware (RTX 4080). The interface shows interactive retrieval of a multi-year phenological trajectory (inset) at a selected pixel in the Amazon.
(c) \textbf{Surface state reconstruction:} A global snapshot of the reconstructed NDVI field on June 24, 2007, showing the large-scale vegetation state during Northern Hemisphere summer.
(d) \textbf{Illustrative temporal derivative field:} An instantaneous temporal derivative field ($\partial \mathrm{NDVI}/\partial t$), computed from the learned neural representation via automatic differentiation. The map shows coherent latitudinal patterns consistent with expected phenological gradients, illustrating a capability inherent to continuous neural representations.}\label{fig:planetary}
\end{figure}

We next evaluate GeoNDC in the planetary-scale regime using the global MODIS MCD43A4 reflectance archive spanning 20 years (2005--2024), with $5\,\mathrm{km}$ spatial resolution and 8-day temporal sampling. This experiment addresses a more stringent question than local reconstruction: can a single neural data cube compactly represent a multi-decadal global archive while remaining directly queryable and scientifically useful? As summarized in Fig.~\ref{fig:planetary}, the experiment validates the framework across three tightly coupled dimensions: substantial storage reduction, direct spatiotemporal queryability, and continuity-aware reconstruction.

The first result concerns storage efficiency. The original scientific archive, stored in double-precision floating-point (float64) GeoTIFF format to preserve numerical fidelity, occupies 168\,GB of disk space. Such a data volume is typical of long-term global EO archives, but in practice it introduces substantial cost in movement, replication, and repeated analysis. In contrast, GeoNDC condenses the entire spatiotemporal volume---including spectral values and validity masks---into a single binary model file of 0.44\,GB. As shown in Fig.~\ref{fig:planetary}(a), this corresponds to a compression ratio of approximately $380\!:\!1$ relative to the float64 raster archive. To provide a stricter comparison, we also considered a more compact raster baseline in which the archive is quantized to Int16 format, reducing the nominal storage volume to approximately 42\,GB. Even against this more conservative reference, GeoNDC still delivers an approximately $95\times$ reduction. These results indicate that the dominant spatiotemporal structure of multi-decadal planetary surface dynamics can be encoded in a neural model that is dramatically smaller than the explicit raster archive.

To quantify reconstruction fidelity at planetary scale, we validated GeoNDC against the original MODIS MCD43A4 observations for the year 2021 (46 temporal frames). Across all seven spectral bands, the mean $R^2$ exceeds 0.98, with individual bands ranging from 0.983 (Band~5, 1240\,nm) to 0.995 (Band~3, Blue). The mean RMSE across bands is 0.021 in reflectance units (scale $[0,1]$), and the mean MAE is 0.011. For the derived NDVI index, the reconstruction achieves a mean $R^2$ of 0.94, a mean RMSE of 0.077, and a mean MAE of 0.031 over the 46 evaluation dates. These accuracy levels are maintained consistently throughout the year without observable seasonal degradation. The quantitative results confirm that, despite the approximately $380\!:\!1$ compression ratio, GeoNDC preserves the dominant spectral and spatiotemporal structure of the original global archive with high fidelity.

To contextualize the compression--fidelity trade-off, we compared GeoNDC against representative baseline methods on four single-frame global snapshots (Table~\ref{tab:compression}). COIN~\cite{dupont2021coin}, a coordinate-based implicit neural representation that independently fits a SIREN MLP to each image, was evaluated at three model sizes. At a comparable per-frame storage budget ($\approx$0.5\,MB), GeoNDC achieves an $R^2$ of 0.994 versus 0.954 for COIN, corresponding to a $2.7\times$ reduction in RMSE (0.020 vs.\ 0.054). This advantage arises because GeoNDC amortizes its parameters across the full temporal extent of the archive, enabling shared spatiotemporal representations that a per-frame model cannot exploit. Traditional lossless and near-lossless methods (Int16 quantization with Zstandard compression, JPEG2000) achieve near-perfect reconstruction ($R^2 \approx 1.0$) but require $60$--$85$\,MB per frame, yielding compression ratios of only $20$--$30\!:\!1$---two orders of magnitude less compact than GeoNDC.

\begin{table}[t]
\centering
\caption{Single-frame compression comparison on global MODIS snapshots (mean over four dates spanning 2021). Per-frame size for GeoNDC is the total model size amortized over all 915 temporal frames.}
\label{tab:compression}
\renewcommand{\arraystretch}{1.1}
\setlength{\tabcolsep}{5pt}
\begin{tabular}{lrrrr}
\hline
\textbf{Method} & \textbf{Size (MB)} & \textbf{Ratio} & $\boldsymbol{R^2}$ & \textbf{RMSE} \\
\hline
GeoNDC (amortized)   & 0.52  & 3\,486:1   & 0.9943 & 0.0199 \\
COIN-L (265k params) & 0.53  & 3\,385:1   & 0.9535 & 0.0543 \\
COIN-M (67k params)  & 0.13  & 13\,360:1  & 0.9445 & 0.0589 \\
COIN-S (9k params)   & 0.02  & 100\,322:1 & 0.9047 & 0.0769 \\
Int16 + Zstandard    & 80.68 & 22:1       & 1.0000 & 0.0000 \\
JPEG2000 (Q=20)      & 61.83 & 30:1       & 1.0000 & 0.0007 \\
\hline
\end{tabular}
\end{table}

Beyond storage reduction, the more consequential property is that the compressed representation remains directly queryable. In conventional raster workflows, retrieving a long-term time series for a single coordinate often requires opening and seeking through thousands of separate files, making the process primarily I/O-bound. GeoNDC changes this access pattern by converting data retrieval into neural function evaluation. Since the entire model fits within the VRAM of a single consumer-grade GPU (e.g., NVIDIA RTX 4080), the framework supports on-demand spatiotemporal access without full decompression of the original archive. As illustrated in the software interface in Fig.~\ref{fig:planetary}(b), users can interactively query arbitrary coordinates on the globe to retrieve and visualize a 20-year phenological trajectory. In this sense, GeoNDC is not merely a compact codec, but a queryable neural data cube: the archive is represented as an executable model rather than only a collection of static rasters.

This queryability extends beyond desktop GPU environments. Because the neural inference pipeline consists only of hash table lookups and lightweight MLP evaluation, it can be executed entirely within a web browser using WebGPU compute shaders. We deployed the global MODIS model as a browser-based viewer at \url{https://www.geondc.org/viewer}, where users can navigate the 20-year record, render arbitrary-time global snapshots, and retrieve per-pixel phenological trajectories without any backend computation or software installation. The 0.44\,GB model file is downloaded once and executed client-side, making decadal planetary analysis accessible from standard laptops and even mobile devices.

Fig.~\ref{fig:planetary}(c) shows the reconstructed NDVI field on June 24, 2007, demonstrating seamless global vegetation mapping during Northern Hemisphere summer. As an illustrative example of the continuous representation, Fig.~\ref{fig:planetary}(d) displays a temporal derivative field ($\partial \mathrm{NDVI}/\partial t$) computed via automatic differentiation, revealing coherent latitudinal phenological gradients.

\begin{figure}[htbp]
\centering
\includegraphics[width=\textwidth]{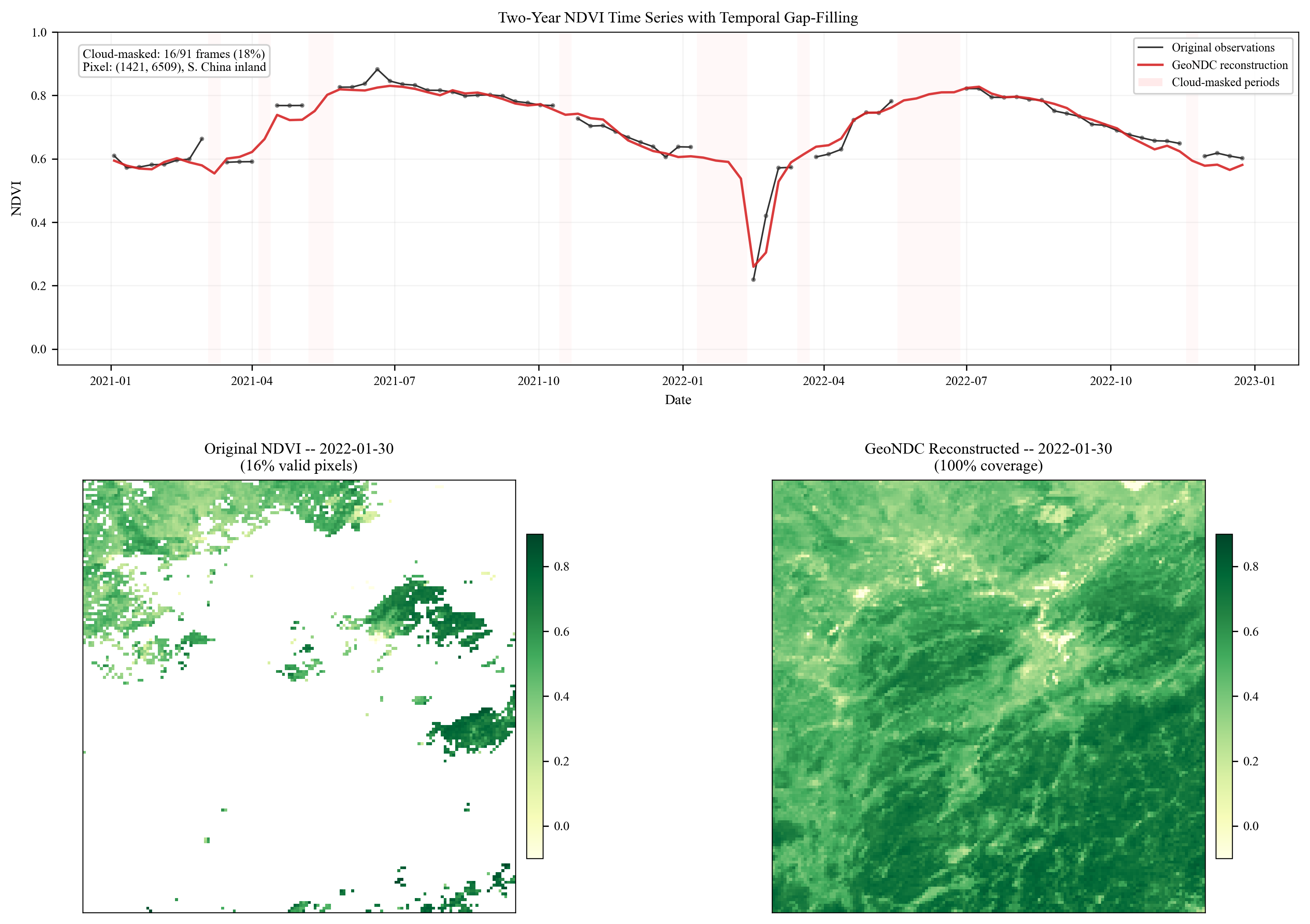}
\caption{\textbf{Temporal gap-filling in a cloud-affected sub-region (southern China).}
\emph{Top:} Two-year NDVI time series at a representative vegetation pixel. Pink shading indicates cloud-masked observation dates. GeoNDC (red curve) closely tracks the original observations (dark line) where data are available and provides continuous reconstruction through cloud-masked gaps. The seasonal phenological cycle is preserved without external gap-filling.
\emph{Bottom:} Spatial comparison on January 30, 2022, when only 16\% of pixels are valid. The original NDVI map (left) shows extensive data loss due to winter cloud cover, while GeoNDC (right) reconstructs a spatially complete vegetation field.}\label{fig:gapfilling}
\end{figure}

The temporal advantage of GeoNDC becomes especially pronounced in regions with persistent cloud cover. To illustrate this, we selected a $300 \times 300$ pixel sub-region over southern China ($\sim$28\textdegree N, 112\textdegree E), where cloud contamination renders 18\% of frames incomplete at representative vegetation pixels over the 2021--2022 period. As shown in Fig.~\ref{fig:gapfilling} (top), GeoNDC reconstructs a continuous, temporally coherent NDVI time series that closely follows the original observations where data are available and smoothly fills cloud-masked gaps without any external gap-filling procedure. The reconstructed curve preserves both the seasonal amplitude and the fine temporal structure of the vegetation phenology. By contrast, per-frame compression methods such as COIN~\cite{dupont2021coin} must train a separate model for every observed snapshot and fundamentally cannot reconstruct frames for which observations are missing. Fig.~\ref{fig:gapfilling} (bottom) further illustrates this on January 30, 2022, when only 16\% of pixels in the sub-region are valid: the original NDVI map shows extensive data loss, while GeoNDC produces a spatially complete and coherent reconstruction by leveraging the temporal context encoded in the shared neural representation.

Overall, the global MODIS experiment shows that GeoNDC can compactly encode a multi-decadal planetary archive, support direct spatiotemporal querying, and provide temporally coherent gap-filling without file-by-file decoding, defining its practical value as a planetary-scale neural data cube.

\FloatBarrier
\subsection{Synergistic Compression of Multi-Product Biophysical Variables}\label{sec:synergistic}

To evaluate whether GeoNDC can serve not only reflectance archives but also analysis-ready ecological products, we applied the framework to the high-resolution ($20\,\mathrm{m}$) HiGLASS Leaf Area Index (LAI) and Fraction of Photosynthetically Active Radiation (FPAR) datasets. In conventional practice, these quantitative inversion products are distributed as disjointed collections of GeoTIFF tiles, which creates substantial storage overhead and repeated I/O friction for long-term ecological monitoring and cross-product analysis.

Rather than compressing LAI and FPAR independently, we adopted a synergistic representation strategy in which both variables are jointly encoded into a single \texttt{.gndc} model. By leveraging the multi-channel output capability of the MLP decoder, the framework encapsulates both variables within a shared spatiotemporal embedding. This design is not only computationally efficient, but also physically meaningful, because LAI and FPAR are strongly coupled through canopy structure and radiation interception. The shared latent representation therefore acts as a biophysically informed regularizer, encouraging the reconstructed products to remain mutually consistent while reducing redundant storage across variables.

\begin{figure}[htbp]
\centering
\includegraphics[width=\textwidth]{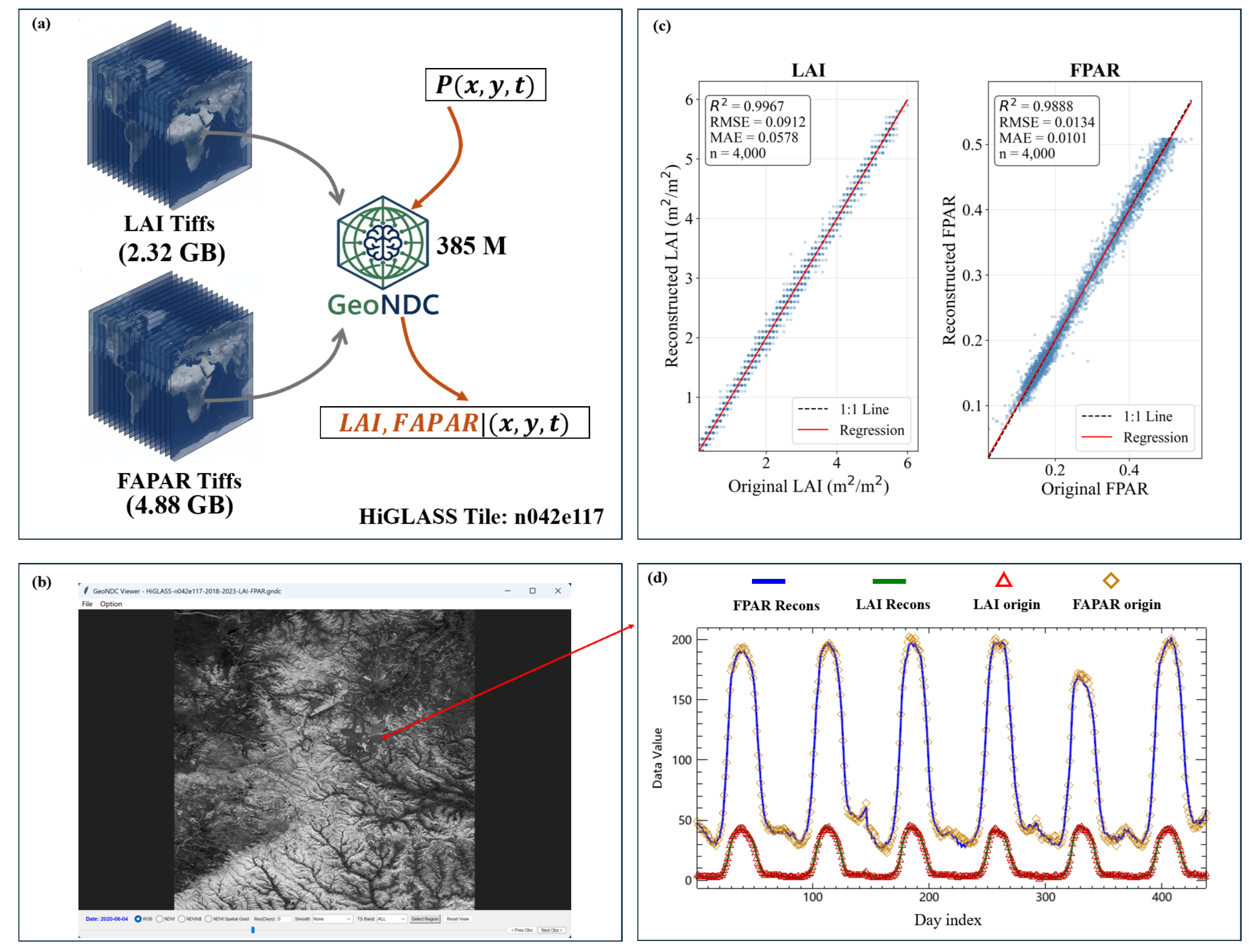}
\caption{\textbf{Shared neural representation of multi-product biophysical variables (LAI and FPAR).} 
(a) \textbf{Joint representation workflow:} Two large discrete GeoTIFF archives (2.32\,GB for LAI and 4.88\,GB for FPAR) are jointly parameterized into a single compact GeoNDC model (385\,MB) through a shared spatiotemporal representation. 
(b) \textbf{Interactive product serving:} The GeoNDC Viewer interface demonstrates direct rendering and exploration of the compressed high-resolution ($20\,\mathrm{m}$) HiGLASS tile. 
(c) \textbf{Quantitative reconstruction accuracy:} Density scatter plots for 4,000 randomly sampled points across diverse seasons show near-perfect agreement between the original and reconstructed values for both LAI ($R^2 = 0.9967$) and FPAR ($R^2 = 0.9888$). 
(d) \textbf{Temporal consistency:} Comparison of multi-year temporal trajectories shows that the reconstructed series preserves the seasonal amplitude, peak vegetation states, and fine temporal oscillations without introducing obvious smoothing artifacts.}\label{fig:synergistic}
\end{figure}

We validated this synergistic framework using 438 temporal observations from the HiGLASS dataset spanning 2018--2023, with a 5-day temporal resolution. To ensure robust evaluation, we randomly sampled approximately 4,000 points across diverse spatial locations and seasonal phenological states. As shown in Fig.~\ref{fig:synergistic}, the reconstructed products exhibit exceptional fidelity to the original data. The reconstructed LAI achieves an $R^2$ of 0.9967, an RMSE of $0.0912\,\mathrm{m}^2/\mathrm{m}^2$, and an MAE of $0.0578\,\mathrm{m}^2/\mathrm{m}^2$. Similarly, the FPAR reconstruction reaches an $R^2$ of 0.9888, an RMSE of 0.0134, and an MAE of 0.0101. These near-perfect correlations indicate that GeoNDC successfully captures the complex spatiotemporal variability of multiple biophysical variables simultaneously, preserving their seasonal dynamics while greatly reducing distribution friction.

The significance of this experiment is not only the reconstruction accuracy, but also the compactness of the joint representation. As illustrated in Fig.~\ref{fig:synergistic}(a), the original HiGLASS products occupy 2.32\,GB for LAI and 4.88\,GB for FPAR, whereas GeoNDC jointly represents them in a single 385\,MB model file. This reduction is especially relevant for ecological applications where multiple products are typically used together, and repeated loading of separate tiled rasters becomes a major bottleneck. The viewer shown in Fig.~\ref{fig:synergistic}(b) further illustrates that the learned representation can be directly rendered and explored as a high-resolution product layer, rather than merely stored as a static archive.

An additional point is that the original HiGLASS products are distributed as quantized integers with inherent scaling factors of 0.1 for LAI and 0.004 for FPAR. In this context, the achieved reconstruction errors are particularly informative. The LAI MAE of 0.0578 falls below the original quantization interval, indicating sub-quantization reconstruction fidelity. For FPAR, the MAE of 0.0101 remains highly constrained in absolute terms and corresponds to approximately 2.5 quantization steps. Thus, despite the substantial storage reduction, the continuous GeoNDC encoding remains effectively lossless for many downstream ecological applications. As shown in Fig.~\ref{fig:synergistic}(d), the reconstructed temporal trajectories also preserve the amplitude and timing of seasonal oscillations without introducing obvious smoothing artifacts.

Taken together, the HiGLASS experiment demonstrates that GeoNDC is not restricted to reflectance archives or visualization-oriented products. It can also serve as a compact and faithful representation for analysis-ready biophysical variables, with a shared latent space that preserves inter-variable consistency while substantially reducing storage and access overhead. This capability is particularly important for future AI-ready EO infrastructures, where multiple scientific products must be jointly queried, compared, and integrated within a unified representational framework.

\FloatBarrier
\subsection{Query Efficiency}\label{sec:query_efficiency}

A core claim of GeoNDC is that the learned neural representation is directly queryable without full decompression of the original archive. To quantify this advantage, we benchmarked the query latency of GeoNDC against conventional GeoTIFF raster access on the 2021 MODIS MCD43A4 archive (46 temporal frames, 4008$\times$8016 pixels $\times$ 7 bands stored as float64 GeoTIFF). Two representative query patterns were evaluated: (1)~a single-pixel time series retrieval across all 46 time steps, and (2)~a regional subset query over a $300 \times 300$ pixel area across all 46 time steps. For the GeoTIFF baseline, each query requires opening and reading from 46 individual raster files stored on a local NVMe SSD. For GeoNDC, all queries are served by neural inference from a single 264\,MB model preloaded into GPU memory. All benchmarks were performed on a workstation equipped with an NVIDIA RTX 4080 GPU (16\,GB VRAM) and a PCIe 4.0 NVMe SSD, with five repetitions per query type.

As shown in Table~\ref{tab:query_efficiency}, GeoNDC achieves substantial speedups for both query types. The most dramatic improvement occurs in single-pixel time series retrieval: GeoNDC returns the complete 20-year time series (all 915 time steps across 7 bands) in approximately 8\,ms, compared to 612\,ms required by conventional raster access to retrieve only the 46 time steps of 2021---an $81\times$ speedup. This advantage arises because GeoTIFF access requires opening 46 separate files and seeking to the target pixel location in each, whereas GeoNDC evaluates the neural field in a single batched GPU forward pass. For regional subset queries ($300 \times 300 \times 46$), GeoNDC achieves a $6.2\times$ speedup (472\,ms vs.\ 2.9\,s).

Beyond query latency, the memory footprint comparison further highlights the practical advantages of the neural representation. Each GeoTIFF frame occupies 1.7\,GB of RAM when loaded (float64, 7 bands), and loading the full 2021 archive (46 frames) would require approximately 77\,GB---exceeding the physical memory of most common desktops. In practice, conventional workflows must therefore process frames sequentially, incurring repeated I/O overhead. In contrast, the GeoNDC model occupies 2.2\,GB of GPU VRAM after loading and provides immediate access to the entire 20-year archive (915 temporal frames), not merely the 46 frames of 2021. This means that GeoNDC delivers faster query performance while simultaneously covering a $20\times$ longer temporal extent within a memory footprint that is smaller than a single GeoTIFF frame.

\begin{table}[t]
\centering
\caption{Query efficiency comparison between GeoNDC and conventional GeoTIFF raster access on the 2021 MODIS MCD43A4 archive (46 temporal frames). GeoNDC serves all queries from a single neural model covering the full 20-year archive (915 frames), preloaded into GPU memory. GeoTIFF queries read from individual float64 raster files stored on a local NVMe SSD.}
\label{tab:query_efficiency}
\renewcommand{\arraystretch}{1.15}
\setlength{\tabcolsep}{5pt}
\begin{tabular}{lrrr}
\hline
\textbf{Query type} & \textbf{GeoTIFF} & \textbf{GeoNDC} & \textbf{Speedup} \\
\hline
Single-pixel time series (1\,px $\times$ 46\,steps)\textsuperscript{a} & 612\,ms  & 8\,ms   & 81$\times$ \\
Regional subset (300$^2$ $\times$ 46\,steps)         & 2.9\,s   & 472\,ms & 6.2$\times$ \\
\hline
\textbf{Per-frame memory (loaded)} & 1.7\,GB RAM (1\,day) & \multicolumn{2}{l}{2.2\,GB VRAM\textsuperscript{b} (20\,yr)} \\
\textbf{On-disk size} & 8.7\,GB (1\,yr) & \multicolumn{2}{l}{264\,MB (20\,yr)} \\
\hline
\multicolumn{4}{l}{\footnotesize \textsuperscript{a}GeoNDC returns all 915 time steps (20\,yr) in 8\,ms; GeoTIFF retrieves only the 46 steps of 2021.} \\
\multicolumn{4}{l}{\footnotesize \textsuperscript{b}Full 20-year model loaded once; provides access to all 915 frames.}
\end{tabular}
\end{table}

\section{Discussion}\label{sec:discussion}

\subsection{Semantic Compression and Topology-Aware Reconstruction}

GeoNDC is designed to complement, rather than replace, authoritative raw observational archives. Its role is to provide a compact, analysis-ready representation layer that reduces the friction between data possession and data use. The approximately $380\!:\!1$ compression ratio on the global MODIS archive and the joint encoding of HiGLASS LAI/FPAR products suggest that the effective dimensionality of many long-term land-surface dynamics is substantially lower than what grid-based storage implies. In practical terms, this makes decadal-scale global analysis feasible on standard local hardware, rather than requiring institutional infrastructures with high-throughput storage.

Importantly, GeoNDC performs more than volumetric compression: it achieves a form of \emph{semantic compression} by organizing scattered, incomplete observations into a queryable continuous representation. This is most evident in missing-data recovery. Unlike traditional preprocessing workflows---such as Maximum Value Compositing or temporal interpolation---that treat cloud removal as a separate heuristic stage, GeoNDC recovers missing values as a natural consequence of fitting a continuous spatiotemporal field. The Sentinel-2 Mask-and-Restore experiment illustrates this clearly: under simulated gaps up to 2\,km in scale, GeoNDC maintained $R^2 > 0.85$ while preserving sharp urban edges and field boundaries, whereas linear interpolation introduced substantial spectral bias in the NIR band during peak growing season. This advantage arises because the model reconstructs missing values conditioned on the learned spatiotemporal topology of the surface process, rather than treating gaps as local holes to be patched independently.

\subsection{Limitations and Future Challenges}

The transition from raster archives to implicit neural representations introduces several trade-offs that should be recognized explicitly.

First, there is a \emph{computational asymmetry}: while the resulting \texttt{.gndc} model is compact and deployable on consumer hardware, the initial encoding process remains computationally intensive. GeoNDC shifts part of the computational burden from repeated downstream I/O toward an upfront learning stage.

Second, the framework exhibits a \emph{representational bias} toward smooth, correlated signals. The dual-branch architecture and optional sparse residual layer mitigate this issue, but extremely transient events---flash floods, short-lived fire fronts, or abrupt disturbances---may still be underrepresented. Whether GeoNDC extends equally well to variables with weaker continuity assumptions or more abrupt regime shifts remains an open question.

Third, GeoNDC is a \emph{lossy representation}. Although reconstruction fidelity is high and in some cases sub-quantization relative to distributed products, the model should still be regarded as an approximation layer. The distinction between raw observations and neural reconstruction---preserved through validity masks and optional correction layers---is a requirement for scientific transparency, not merely an implementation detail.

These limitations clarify the regime in which GeoNDC is most useful: long-term, high-volume, strongly structured EO archives where repeated access and continuity-aware reconstruction are central needs. Future improvements may include adaptive hash allocations, uncertainty-aware neural data cubes, derivative-regularized encoding architectures, event-sensitive residual schemes, and tighter integration with cloud-native geospatial infrastructures.

\subsection{Outlook}

The broader implication of this work is that planetary EO archives may be amenable to neural parameterization when their dominant dynamics can be represented as structured spatiotemporal fields. While the present study focuses on land-surface reflectance and biophysical variables, the underlying principle extends to other classes of geospatial archives. The transition from \emph{data as files} to \emph{data as models} opens the possibility of a more interactive, analysis-ready planetary data infrastructure in which storage, access, and reconstruction become properties of the representation itself. Realizing this vision will require further advances in uncertainty quantification, multi-sensor harmonization, dynamic updating, and scalable training strategies.

\section{Conclusion}\label{sec:conclusion}

In this work, we introduced GeoNDC, a queryable neural data cube for planetary-scale Earth observation. By reformulating georeferenced EO archives as continuous spatiotemporal neural representations, GeoNDC unifies compact storage, direct spatiotemporal query, and continuity-aware reconstruction within a single framework. Experiments on Sentinel-2, global MODIS, and high-resolution HiGLASS products show that this representation preserves fine surface structure under incomplete observations, compresses a 20-year global archive from 42\,GB (Int16) to 0.44\,GB while retaining the dominant spatiotemporal dynamics, and jointly represents strongly coupled biophysical variables with near-lossless fidelity. Beyond compression, the main contribution of GeoNDC is representational: the same learned model supports query and reconstruction from a unified neural data cube. As a lossy neural representation, GeoNDC is not intended to replace authoritative raw EO archives, but to complement them with a compact, analysis-ready and AI-ready layer for downstream use. More broadly, GeoNDC points toward a shift from \emph{data as files} to \emph{data as models}, in which planetary-scale EO archives can be represented as compact, queryable computational objects.

\subsection*{Acknowledgments}
This work was supported by the Beijing Nova Program (20240484556) and National Natural Science Foundation of China (Grant No. 42192580), .

\subsection*{Data and Code Availability}
The source code for GeoNDC is publicly available at \url{https://github.com/jianboqi/pygndc}. An interactive browser-based viewer for exploring GeoNDC models, including sample data used in this study, is available at \url{https://www.geondc.org}.

\bibliography{references}

\end{document}